\documentclass{article}
\usepackage{ijcai19}

\usepackage{times}
\usepackage{soul}
\usepackage{url}
\usepackage[hidelinks]{hyperref}
\usepackage[utf8]{inputenc}
\usepackage[small]{caption}
\usepackage{graphicx}
\usepackage{amsmath}
\usepackage{booktabs}
\usepackage{algorithm}
\usepackage{algorithmic}
\usepackage{pgfplots}
\usepackage{svg}
\usepackage{multirow}
\usepackage[linguistics]{forest}
\title{Improving Cross-Domain Performance for Relation Extraction via \\ Dependency Prediction and Information Flow Control}

\author{
Amir Pouran Ben Veyseh,
Thien Nguyen \And
Dejing Dou
\affiliations
Department of Computer and Information Science, University of Oregon, OR, USA\\
\emails
\{ apouranb, thien, dou\}@cs.uoregon.edu
}

\begin{document}

\maketitle

\begin{abstract}
Relation Extraction (RE) is one of the fundamental tasks in Information Extraction and Natural Language Processing. Dependency trees have been shown to be a very useful source of information for this task. The current deep learning models for relation extraction has mainly exploited this dependency information by guiding their computation along the structures of the dependency trees. One potential problem with this approach is it might prevent the models from capturing important context information beyond syntactic structures and cause the poor cross-domain generalization. This paper introduces a novel method to use dependency trees in RE for deep learning models that jointly predicts dependency and semantics relations. We also propose a new mechanism to control the information flow in the model based on the input entity mentions. Our extensive experiments on benchmark datasets show that the proposed model outperforms the existing methods for RE significantly.
\end{abstract}

\section{Introduction}


Extracting semantic relations between entity pairs in text (i.e., Relation Extraction (RE)) is an important task of information extraction. In this paper, we focus on the usual single-sentence setting where two entity mentions appear in the same sentence and the goal is to identify their semantic relationship within this sentence. 
RE has a wide range of downstream applications, including question answering and knowledge base population.


Among many different approaches, deep learning has proven itself as a very effective method for RE in recent research \cite{xu2015classifying,nguyen2015combining,Wang:16,fu2017domain,shi2018genre}. The major factors that contribute to the success of deep learning for RE involve pre-trained word embeddings to generalize words and deep learning architectures to compose the word embeddings to induce effective representations. Recently, dependency trees have also been shown to be useful for deep learning models applied to RE \cite{miwa2016end,nguyen2015combining,zhang2018graph}. The typical way to exploit dependency trees for RE in deep learning models is to rely on the dependency structures to guide the computations of the models. In particular, the shortest dependency paths between the two entity mentions have been exploited to form sequences of words for Long Short-Term Memory (LSTM) networks \cite{Xu:15} while the dependency trees themselves are employed to direct the operation of the graph-based convolutions in recent deep learning models (i.e., Graph Convolutional Neural Networks \cite{zhang2018graph})


Despite the good performance of these methods to exploit dependency information, there are at least two issues that might prevent them from further improving the performance. First, as the information flow in the models is restricted to the structures of the trees, the direct application of the dependency trees in the models might fail to capture important context information that goes beyond the coverage of such tree structures. Second, in the cross-domain setting where the sentences in the training data and test data come from different domains, the dependency structures in the training data might also be dissimilar to those in test data. If a model is trained with the structure guidance for the training data, it might not be able to generalize to the structure in the test data, causing the poor performance in the cross-domain setting.


%

%


In order to overcome the aforementioned issues, we introduce a novel method to exploit dependency trees for RE in which the dependency structures of the sentences serve as the ground-truth in a multi-task learning setting to perform both RE and dependency relation prediction simultaneously. In this way, instead of using the dependency trees directly as in the previous approaches, we are using the dependency structures indirectly to encourage the induced representations to be more general with respect to both semantic and dependency relations. Specifically, we first use a modified version of self-attention \cite{Vaswani:17} to learn a context-aware representation for each token in the sentence. The self-attention representation for each word would encode rich semantic structures, reflecting the semantic portion that one word would contribute to the others in the sentences. Note that such pairwise semantic contributions are crucial to the semantic prediction in RE. Given the representations of words from the self-attention module, we jointly predict the dependency relations between every pair of words in the sentences, causing the word representations to encapsulate both the semantic and syntactic structures of the sentences. Technically, if we build a semantic adjacency matrix for the words in the sentences based on the pairwise similarities with the self-attention representations of the words, the dependency relation predictions would regulate this semantic adjacency matrix to be similar to the syntactic adjacency matrix induced by the dependency trees. On the one hand, as the semantic self-attention representations of the words have the context information of the whole sentences, once used to make RE prediction, it might help to capture the important context information that the dependency structures cannot reach. On the other hand, the breakdown of the dependency trees into dependency edges/relations in the prediction framework eliminates the emphasis on the whole tree structures that are specific to domains, and focuses on the dependency relations that are shared across domains. This might help to improve the cross-domain performance of the models. Finally, in order to customize the word representations for RE, we propose a novel control mechanism to remove the information that are irrelevant to the semantic prediction of the two entity mentions of interest for RE. In particular, the representations of the two entity mentions are used to compute a semantic control vector that is then applied to the representations of the individual words as a way to retain the most RE-specific information. 



Our experiments on the ACE 2005 dataset shows that our model is able to achieve the state-of-the-art performance in the standard datasets for RE. To summarize, our contributions include:

\begin{itemize}
\item We introduce a novel method to exploit dependency trees for RE with deep learning based on the predictions of dependency relations.
\item We present a novel control mechanism over the feature representations of each word in the sentences to customize the representations for RE.
\item We conduct extensive experiments on benchmark datasets and analyze the performance of the model in cross-domain relation extraction.
\end{itemize}

\section{Model}


The RE problem in this work is defined as follows: given an input sentence $X = x_1,x_2,\ldots,x_n$ ($x_i$ is the $i$-th word in the sentence) and the two indexes $s$ and $o$ for the two entity mentions of interest (called relation mention), we would like to predict the semantic relationship between these two mentions. If there is no relation between $x_s$ and $x_o$, we assign label $None$ for the relation mention. There are three major components in the model proposed in work: (1) Representation Learning: to learn a feature representation for each word based on the semantic and dependency structures of the sentences, (2) Representation Controlling: to determine which features for each token should be used in the final representation based on the two entity mention of interest, and finally (3) to predict the semantic relation for two entity mentions based on the learned representations of the tokens. In the following we describe each part in detail.

\subsection{Representation Learning}
\label{sec:rl}

In order to prepare the input sentence for the neural computation in the following steps, we first transform each word in $X$ into a real-valued representation vector. Inspired by the previous work on relation extraction \cite{nguyen2015combining,fu2017domain}, we use vector $w_i = [e_i,ps_i,po_i,t_i,c_i,p_i,g_i]$ to present each word $x_i \in X$ where:
\begin{itemize}
    \item $e_i$ is some pre-trained word embedding $x_i$.
    \item $ps_i$ and $po_i$ are position embedding vectors to indicate the relative distances from the current token $x_i$ to the two entity mentions of interest $x_s$ and $x_o$ (i.e., $i-s$ and $i-o$) respectively.
    \item $t_i$ and $c_i$ are embedding vectors for the tags of $x_i$ to reflect the entity mention and chunk information in $X$ (following the BIO tagging schema) respectively.
    \item $p_i$ is a binary number that is 1 if $x_i$ is on the dependency path between $x_s$ and $x_o$ in the dependency tree of $X$; otherwise it is zero.
    \item $g_i$ is a binary vector whose size is the total number of dependency relations in dependency trees. The dimension that corresponds to the dependency relation $r$ is set to 1 if $x_i$ has the relation $r$ with some other word in $X$; and 0 otherwise.
\end{itemize}


\subsubsection{Self-Attention Representation}

This word-vector transformation converts the input sentence $X$ into a sequence of representation vectors $W = w_1,w_2,\ldots,w_n$ for the words in $X$. In this vector sequence, $w_i$ only encapsulates information about the token $x_i$ itself. In order to encode richer context information of the whole sentence into the representations for each word in $X$, we run a bidirectional LSTM network over $W$, generating the sequence of hidden vectors $h_1,h_2,\ldots,h_n$. Each hidden vector $h_i$ is the concatenation of the corresponding hidden vectors from the forward and backward LSTM networks that compresses the whole information content of $X$ with a greater focus on $x_i$. However, for RE, the context information of $x_s$ tends to be less pronounced in $h_o$ (and vice verse) if $x_o$ is far away from $x_s$ in the sentence due to the gated and recurrent mechanisms with the forget gate of LSTM. This is undesirable as the context information of $x_s$ (or $x_o$) might provide important context information for $h_o$ (or $h_s$) when it comes to predict the semantic relation between $x_s$ and $x_o$. For example, the context information of $x_s$ might help to reveal its entity subtype that once integrated well into $h_o$, can promote $h_o$ as a rich features for the semantic prediction. In order to overcome this issue, we employ the self-attention mechanism that allows a word to directly contribute its context information into the hidden vector of another word only based on the potential semantic contribution, ignoring the distance barriers between the words. Specifically, in the self-attention mechanism, we compute three new vectors $k_i$ (key vector), $q_i$ (query vector) and $v_i$ (value vector) for each token $x_i$ from its hidden vector $h_i$:
\begin{equation}
\begin{split}
    k_i & = W_k * h_i \\
    q_i & = W_q * h_i \\
    v_i & = W_v * h_i \\
\end{split}
\end{equation}
where $*$ is the matrix multiplication operation. Note that for simplicity, we omit the biases in the formula for this paper. Afterward, the potential context contribution of $x_j$ for $h_i$ is determined via the similarity between the the key vector $k_j$ of $x_j$ and the query vector of $x_i$ (i.e., the dot product):
\begin{equation}
    w_{ij} = \frac{\text{exp}(q_i \cdot k_j)}{\sum_{t=1}^n \text{exp}(q_i \cdot k_t)}
\end{equation}
Given these contribution weights, the self-attention representation vectors for the words in the sentence $X$ is generated by the weighted sums of the value vectors:
\begin{equation}
    h'_i = \Sigma_{j=1}^n w_{ij} v_j
\end{equation}


\subsubsection{Dependency Relation Prediction}

The self-attention mechanism helps the representation vectors $h'_i$ to capture the semantic features of the input sentence $X$. This section aims to enrich the vectors $h'_i$ with the syntactic structure of $X$ (i.e., the dependency tree) that has been shown to be helpful for deep learning models for RE. 
As mentioned in the introduction, the traditional methods to use dependency trees to guide the computation of deep learning models would limit the context coverage of the models and cause the poor generalization over dependency structures across domains. In order to avoid such issues, instead of using the dependency trees directly, we break the dependency trees into dependency relations between words that are then employed as the ground-truths to be predicted by the model in a multi-task learning framework for RE. The structure decomposition would help to circumvent the modeling of the whole tree structures to improve the cross-domain generalization while still injecting the syntactic information into the representation vectors via the dependency prediction. In particular, given two words $x_i$ and $x_j$ in the sentence, we first compute the probability $\hat{a}_{i,j}$ to indicate whether $x_i$ and $x_j$ are connected to each other in the dependency tree based on their self-attention representation vectors $h'_i$ and $h'_j$:
\begin{equation}
\label{predict_eq}
    \hat{a}_{i,j} = \text{sigmoid}(W_{d2} * g(W_{d1} * [h'_i,h'_j]))
\end{equation}
where $[u,v]$ is the concatenation operation for the two vectors $u$ and $v$, $W_{d1}$ and $W_{d2}$ are the model parameters and $g$ is a non-linear function. These probabilities are then employed in a loss function to maximize the likelihood of the dependency connections in the dependency tree:
\begin{equation}
\begin{split}
     L_{dep-pred} & = \frac{1}{T} \Sigma_{t=1}^T \Sigma_{i=1}^{n} \Sigma_{j=1}^{n} a_{i,j} \log(\hat{a}_{i,j}) \\ & + (1-a_{i,j})\log(1-\hat{a}_{i,j})
\end{split}
\end{equation}
where $a_{i,j} = 1$ if there is an edge between tokens $x_i$ and $x_j$ in the dependency tree of $X$; and 0 otherwise, and $T$ is the number of examples in the training data. Note that our method of predicting dependency relations to enrich word representations for RE is similar to the method employed by \cite{strubell2018linguistically} for another task of semantic role labeling. However, in the proposed method we predict the existence of the dependency edges between every pair of words in the sentence while \cite{strubell2018linguistically} only predict the dependency heads of the words in the sentence, just only considering the connected pairs of words in the dependency trees and ignoring the other word pairs of the sentence. In the experiments we found that considering the dependency relations for every pair of words is also important for RE.

\subsection{Control Mechanism}

In addition to the indirect use of dependency trees, we introduce a new control mechanism for RE that regulates the information flow in the model based on the entity mentions of interest. The rationale for this control mechanism is twofold: (1) for RE, the two entity mentions $x_s$ and $s_o$ are crucial and the effective word representations for this task should retain only the relevant information/features with respect to these two entity mentions. The control mechanism functions as the relevant information filter for RE in this work, and (2) in the attention mechanism we compute a single weight for each word, thus assuming the same weights for every dimension/feature in the word's representation vector. In practice, it might be more flexible if we can regulate the individual dimension/feature so the important dimension/feature for RE would be promoted in the representation vectors. The control mechanism would help to quantify the contribution of each dimension/feature to achieve such flexibility.

The model description so far has introduced two types of representation vectors for the words in the sentence, i.e., the initial contextualized word vectors $H = h_1,h_2,\ldots,h_n$ (i.e., the outputs of the bidirectional LSTM layers) and the semantically and syntactically enriched vectors $H' = h'_1,h'_2,\ldots,h'_n$. With the idea of the control mechanism, we seek to manipulate the word representations in both $H$ and $H'$ at the feature level so the resulting representation vectors would be specific to the two entity mentions $x_s$ and $x_o$. In particular, we start with the initial contextualized word vectors in $H$ where the hidden vectors $h_s$ and $h_o$ for $x_s$ and $x_o$ are used to generate the control vector $p$ for $H$ via:
\begin{equation}
    p = Relu (W_p * [h_s,h_o])
\end{equation}
Given the control vector $p$, we filter the irrelevant information (with respect to $x_s$ and $x_o$) in the representation vectors of $H$ via the element-wise multiplication $\odot$, transforming each vector $h_i \in H$ into the filtered vector $\bar{h}_i$:
\begin{equation}
    \bar{h}_i = p \odot h_i
\end{equation}
Note that the element-wise multiplication operation allows us to control the representation vectors at the feature level. In the next step, we compute the control vector $c$ for the vectors in $H'$ based on two sources of information specific to $x_s$ and $x_o$: (i) the initial contextualized vectors $h_s$ and $h_o$ for $x_s$ and $x_o$ (as does for the vectors in $H$), and (ii) the weighted sum of the vectors in $H$. In order to generate the weight for each vector in $H$, we rely on the filtered vectors $\bar{h}_i$ to ensure that the weights are customized for two entity mention of interest:
\begin{equation}
\begin{split}
    \alpha_i & = \frac{\text{exp}(W_{\alpha} \bar{h}_i)}{\sum_{j=1}^n\text{exp}(W_{\alpha} \bar{h}_j)} \\
    m  = \sum_{i=1}^n \alpha_i h_i &,\text{  }  c = Relu(W_c [m, h_s, h_o])
\end{split}
\end{equation}
The control vector $c$ is then applied to each self-attention vector in $h'_i \in H'$ to produce the final representation vector $\bar{h}'_i$  (via the element-wise multiplication $\odot$) that is both specialized for the two entity mentions, and semantically and syntactically enriched for RE:
\begin{equation}
    \bar{h}'_i = c \odot h'_i
\end{equation}


\subsection{Prediction}
\label{sec:pred}

In the prediction step, we utilize the induced representation vectors in the previous steps to perform the relation prediction for $x_s$ and $x_o$ in $X$. In particular, following \cite{zhang2018graph}, we use the following aggregation vector $o$ as the features for the final prediction:
\begin{equation}
    o = [h_s,h_o,\bar{h}'_s,\bar{h}'_o, \text{max}(\bar{h}'_1,\bar{h}'_2,\ldots,\bar{h}'_n)]
\end{equation}
where $\text{max}(\bar{h}'_1,\bar{h}'_2,\ldots,\bar{h}'_n)$ is the element-wise max operation that retains the highest values along the dimensions of the vectors. Note that the vectors in $o$ capture the context information for the $x_s$ and $x_o$ in $X$ at different levels of abstraction to improve the representatives of the features for RE. In particular, $h_s$ and $h_o$ encode the initial contextualized representation at the basic level while $\bar{h}'_s,\bar{h}'_o$ involve a deeper abstraction level with semantic, syntactic and customized features at the two entity mentions. $\text{max}(\bar{h}'_1,\bar{h}'_2,\ldots,\bar{h}'_n)$ goes one step further to select the most important rich features across the whole sentence. For prediction, the feature vector $o$ would be fed into a two-layer feed forward neural network followed by a softmax layer in the end to compute the probability distribution $P_y$ over the possible relation labels for RE:
\begin{equation}
    P(.|X,s,o) = \text{softmax}(W_2 * (W_1 * o))
\end{equation}
We employ the negative log-likelihood as the loss function for the relation prediction in this work:
\begin{equation}
     L_{label} = -\log P(y|X,s,o)
\end{equation}
where $y$ is the correct relation label for $x_s$ and $x_o$ in $X$. Overall, we optimize the following combined loss function $L$ for the model parameters:
\begin{equation}
     L = L_{label} + \lambda L_{dep-pred}
\end{equation}
 where $\lambda$ is the trade-off parameter between the losses for relation prediction and dependency relation prediction we discussed above.
 
 Finally, in order to update the parameters, we use the Adam optimizer with shuffled mini-batches and back-propagation to compute the gradients.

\section{Experiments}
\subsection{Dataset and Parameters}


We evaluate the models in this work using two benchmark datasets for RE, i.e., the ACE 2005 dataset \cite{yu2015combining} and the SemEval 2010 Task 8 dataset \cite{Hendrickx:10}.

For the ACE 2005 dataset, following the previous work \cite{nguyen2015combining,fu2017domain,shi2018genre}, we use the dataset preprocessed and provided by \cite{yu2015combining} for compatible comparison. The ACE 2005 dataset has 6 different domains: broadcast conversation ({\tt bc}), broadcast news ({\tt bn}), conversational telephone conversation ({\tt cts}), newswire ({\tt nw}), usenet ({\tt un}), and webblogs ({\tt wl}). Similar to the prior work, we use the union of the domains {\tt bn} and {\tt nw} (called {\tt news}) as the training data (with 43497 examples) (called the source domain), a half of the documents in {\tt bc} as the development data, and the remainder ({\tt cts}, {\tt wl} and the other half of {\tt bc}) as the test data (called the target domains). Note that we also use the entity mentions, chunks, and dependency trees provided by \cite{yu2015combining} as in the previous work to generate the input features for the words in the sentences. An advantage of the ACE 2005 dataset is it helps to evaluate the cross-domain generalization of the models as the training data and test data in this case comes from different domains.

For the SemEval 2010 Task 8 dataset \cite{Hendrickx:10}, it comes with 9 directed relations with a special class of {\it Other}, leading to a 19-class classification problem. 
As SemEval 2010 does not provide validation data, we use the same model parameters as those used for the ACE 2005 dataset to make it more consistent.
We use the official evaluation script for this dataset to obtain the performance of the models as in the prior work \cite{nguyen2015combining,miwa2016end}.

We fine tune the model parameters on the validation data of the ACE 2005 dataset. The parameters we found include: 50 dimensions for position embedding vectors, the entity mention tag vectors and the chunk tag embedding vectors; 100 hidden units for the bidirectional LSTM network in the representation learning component; 200 dimensions for all the hidden vectors in the model;
 and 0.3 for the learning rate; 0.01 for the trade-off $\lambda$ in the overall loss function. Finally, we use the pre-trained word embedding {\tt word2vec} to initialize the models.


\subsection{Experiments on the ACE 2005 Dataset}

Table \ref{sota-full} compares the proposed model (called {\it DRPC} -- Dependency Relation Prediction and Control) with the best reported models on the ACE 2005 dataset in the cross-domain setting for RE. Such best reported models include the Factor-based Compositional Embedding Models ({\it FCM}) in \cite{yu2015combining}, the deep learning models (i.e., {\it CNN}, {\it Bi-GRU}) in \cite{nguyen2015combining}, the domain adversarial neural network (i.e., {\it CNN+DANN}) in \cite{fu2017domain} and the current best model with the genre separation network ({\it GSN}) in \cite{shi2018genre}. 


\begin{table}[t!]
\addtolength{\abovecaptionskip}{-2.0mm}
\addtolength{\belowcaptionskip}{-3mm}
\small
\begin{center}
\begin{tabular}{|l|c|c|c|c|}
\hline \textbf{System} & \textbf{bc} & \textbf{cts} & \textbf{wl} & \textbf{Avg.} \\ \hline
FCM \shortcite{yu2015combining} & 61.90 & 52.93 & 50.36 & 55.06 \\
Hybrid FCM \shortcite{yu2015combining} & 63.48 & 56.12 & 55.17 & 58.25 \\
LRFCM \shortcite{yu2015combining} & 59.40 & - & -& - \\
Log-linear \shortcite{nguyen2015combining} & 57.83 & 53.14 & 53.06 & 54.67 \\
CNN \shortcite{nguyen2015combining} & 63.26 & 55.63 & 53.91 & 57.60 \\
Bi-GRU \shortcite{nguyen2015combining} & 63.07 & 56.47 & 53.65 & 57.73 \\
Forward GRU \shortcite{nguyen2015combining} & 61.44 & 54.93 & 55.10 & 57.15\\
Backward GRU \shortcite{nguyen2015combining} & 60.82 & 56.03 & 51.78 & 56.21 \\
CNN+DANN \shortcite{fu2017domain} & 65.16 & - & - & - \\ 
GSN \shortcite{shi2018genre} & 66.38 & 57.92 & 56.84 & 60.38 \\ \hline
DRPC & \textbf{67.30} & \textbf{64.28} & \textbf{60.19} & \textbf{63.92} \\
\hline
\end{tabular}
\end{center}
\caption{\label{sota-full} F1 scores of the models on the ACE 2005 dataset over different target domains {\tt bc}, {\tt cts}, and {\tt wl}.}
\end{table}

As we can see from the table, the proposed model {\it DRPC} is significantly better than all the previous models on the cross-domain setting for ACE 2005 over different target domains {\tt bc}, {\tt cts} and {\tt wl} ($p < 0.05$). This is remarkable as {\it DRPC} does not apply any specific techniques to bridge the gap between domains while the previous work relies on such techniques to be able to perform well across domains (i.e., \cite{fu2017domain} and \cite{shi2018genre} with the domain adversarial training). Such performance improvement of {\it DRPC} demonstrates the effectiveness of the proposed model with self-attention, dependency connection prediction and information flow control in this work.

\begin{table}[t!]
\addtolength{\abovecaptionskip}{-2.0mm}
\addtolength{\belowcaptionskip}{-3mm}
\small
\begin{center}
\begin{tabular}{|l|c|c|c|c|}
\hline \textbf{System} & \textbf{bc} & \textbf{cts} & \textbf{wl} & \textbf{Avg.} \\ \hline
CNN \shortcite{nguyen2015combining} & 46.3 & 40.8 & 35.8 & 40.9 \\
GRU \shortcite{nguyen2015combining} & 45.2 & 40.2 & 35.1 & 40.1 \\
Bi-GRU \shortcite{nguyen2015combining} & 46.7 & 41.2 & 36.5 & 41.4 \\
GSN \shortcite{shi2018genre} & 52.8 & 45.3 & 39.4 & 45.8 \\ \hline
DRPC & \textbf{59.81} & \textbf{57.82} & \textbf{51.24} & \textbf{56.29} \\
\hline
\end{tabular}
\end{center}
\caption{\label{sota-noling} Performance on the ACE 2005 test sets when linguistic features are not used.}
\end{table}

In order to further evaluate the models, Table \ref{sota-noling} reports the performance of the models when the linguistic features for the input vectors in Section \ref{sec:rl} are not included. In particular, we do not use the embedding vectors $t_i$ and $c_i$ for the entity mention and chunk information, and the $p_i$ and $g_i$ features for the dependency trees in this experiment (i.e., only the word embeddings and the position embeddings are kept). It is clear from Tables \ref{sota-full} and \ref{sota-noling} that the performance of the models drops significantly when the linguistic features are excluded. However, the performance of the proposed model {\it DRPC} still significantly outperform the compared models with large performance gap (an absolute F1 improvement of 7.9\%, 18.9\% and 17.0\% over the state-of-the-art model {\it GSN} \cite{shi2018genre} for the domains {\it bc}, {\it cts} and {\it wl} respectively).This helps to further testify to the effectiveness of {\it DRPC} for RE.


\subsection{Comparing to Dependency-based Models}

The previous section has compared {\it DRPC} with the state-of-the-art models on the ACE 2005 dataset. This section focuses on the comparison of {\it DRPC} with the state-of-the-art deep learning models for RE that employ dependency trees in their operation. We perform such comparisons on both the ACE 2005 and SemEval 2018 datasets.

For RE, the best deep learning model with dependency trees is currently the graph convolutional neural network model (i.e., {\it C-GCN}) in \cite{zhang2018graph} where the dependency trees are used to guide the convolutional operations over the input sentences. We use the implementation of {\it C-GCN} provided by \cite{zhang2018graph} and evaluate its performance on the ACE 2005 dataset with the cross-domain setting. In addition, we implement the Linguistically-Informed Self-Attention model ({\it LISA}) in \cite{strubell2018linguistically} and adapt it to our RE problem for evaluation purpose. Note that although {\it LISA} is originally designed for semantic role labeling, not for RE, it represents a recently proposed method to exploit dependency trees in deep learning models to predict relations between two words in a sentence with good performance, thus being eligible for a baseline for our work. {\it C-GCN} and {\it LISA} only involve the use of dependency trees and do not include the control mechanism as we do in this work. For a fairer comparison, we integrate the control mechanism into such models (leading to {\it C-GCN} + Control, and {\it LISA} + Control) and compare them with the proposed model {\it DRPC} in this work. Table \ref{sota-general} presents the performance of the models on the ACE 2005 dataset.


\begin{table}[t!]
\small
\begin{center}
\begin{tabular}{|l|c|c|c|c|}

\hline \textbf{System} & \textbf{bc} & \textbf{cts} & \textbf{wl} & \textbf{Avg.} \\ \hline
C-GCN \shortcite{zhang2018graph} & 62.55 & 62.98 & 55.91 & 59.24 \\
LISA \shortcite{strubell2018linguistically} & 65.04 & 63.21 & 55.18 & 60.13 \\
C-GCN + control & 65.68 & 63.91 & 58.94 & 62.29 \\
LISA + control & 66.36 & 63.39 & 59.17 & 62.78 \\ \hline
DRPC & \textbf{67.30} & \textbf{64.28} & \textbf{60.19} & \textbf{63.92} \\

\hline
\end{tabular}
\end{center}
\caption{\label{sota-general} Model's performance on the ACE 2005 test datasets.}
\end{table}

The results from the table show that the control mechanism is very useful for RE as it helps to improve the performance for both {\it C-GCN} and {\it LISA} over all the three target domains. The improvements are significant except for {\it LISA} on the {\tt cts} domain. More importantly, we see that {\it DRPC} is significantly better than all the compared models over all the target domains with $p < 0.05$, clearly proving the benefits of the dependency relation prediction proposed in this work.

Finally, Table \ref{semEval} compares {\it DRPC} with {\it C-GCN}, {\it LISA}, and the previous dependency-based methods for RE on the SemEval 2010 dataset. We select the dependency-based models reported in \cite{zhang2018graph} as the baselines, including SVM \cite{Hendrickx:10}, {\it SDP-LSTM} \cite{xu2015classifying}, {\it SPTree} \cite{miwa2016end}, and {\it PA-Tree} \cite{zhang2017position}. The table confirms the effectiveness of {\it DRPC} that significantly outperforms all the compared methods.


\begin{table}[t!]
\small
\begin{center}
\begin{tabular}{|l|l|}
\hline \textbf{System} & \textbf{F1} \\ \hline
SVM \shortcite{Hendrickx:10} & 82.2 \\
SDP-LSTM \shortcite{xu2015classifying} & 83.7 \\
SPTree \shortcite{miwa2016end} & 84.4 \\
PA-Tree \shortcite{zhang2017position} & 82.7 \\
C-GCN \shortcite{zhang2018graph} & 84.8 \\
LISA \shortcite{strubell2018linguistically} & 83.9 \\
\hline
DRPC & \textbf{85.2} \\
\hline
\end{tabular}
\end{center}
\caption{\label{semEval} Performance on the SemEval 2010 dataset.}
\end{table}

\subsection{Ablation Study}

Three important components in the proposed model {\it DRPC} include the self-attention layer (called {\it SA}), the dependency relation prediction technique (called {\it DP}), and the control mechanism (called {\it CM}). In order to evaluate the contribution of these components into the model performance, Table \ref{ablation} presents the performance of {\it DRPC} on the ACE 2005 development dataset when such components are excluded one by one from the model. From the table, we can conclude that all the three components {\it SA}, {\it DP} and {\it CM} are important for {\it DRPC} as removing any of them would further decrease the performance of the model.



\begin{table}[t!]
\addtolength{\abovecaptionskip}{-2.0mm}
\addtolength{\belowcaptionskip}{-3mm}
\small
\begin{center}
\begin{tabular}{|l|c|c|c|}
\hline \textbf{System} & \textbf{Precision} & \textbf{Recall} & \textbf{F1} \\ \hline

DRPC & 72.10 & 63.49 & 67.52 \\ \hline \hline
- CM & 74.92 & 60.15 & 66.88 \\
- DP - CM & 71.05 & 62.00 & 64.72 \\
- SA - DP - CM & 69.02 & 57.14 & 61.96 \\

\hline
\end{tabular}
\end{center}
\caption{\label{ablation} Ablation Study. Model's performance on the ACE 2005 developmet set.}
\end{table}

\begin{table}[t!]
\small
\begin{center}
\begin{tabular}{|l|c|c|c|c|c|}
\hline & \textbf{CNN} & \textbf{RNN} & \textbf{C-GCN} & \textbf{LISA} & \textbf{DRPC} \\ \hline
100 \% & 61.95 & 62.08 & 64.28 & 66.72 & 67.52 \\
90 \% & 61.75 & 61.83 & 62.39  & 65.48 & 66.01 \\
80 \% & 56.91 & 57.92 & 58.04  & 61.74 & 63.92 \\
70 \% & 52.87 & 51.05 & 52.74  & 56.81 & 56.98 \\ 
60 \% & 51.35 & 48.32 & 45.91  & 49.92 & 60.39 \\ 
50 \% & 51.34 & 40.55 & 43.39  & 42.84 & 57.00 \\ 
40 \% & 44.53 & 41.73 & 36.17  & 41.50 & 56.93 \\ 
30 \% & 31.45 & 32.52 & 28.14  & 33.86 & 50.49 \\ 
20 \% & 21.13 & 22.48 & 24.66  & 26.75 & 41.59 \\ 
10 \% & 20.85 & 19.03 & 19.08  & 21.91 & 26.69 \\ 
\hline
\end{tabular}
\end{center}
\caption{\label{complexity} Complexity analysis of the models. The first columns show how much of training data has been used for training. Performance is on the ACE 2005 development set.}
\end{table}

\subsection{Analysis \& Discussion}


In this section, we first evaluate the sample complexity of the models to better understand their operation. In particular, we choose different subsets of the ACE 2005 training dataset according to different ratios of the size (i.e., 10\%, 20\%, 30\% etc.). Such subsets are then used to train the models to evaluate their performance. Table \ref{complexity} shows the performance of {\it DRPC} and 4 other baselines, including {\it CNN} and {\it Bi-GRU} in \cite{nguyen2015combining}, {\it C-GCN} \cite{zhang2018graph} and {\it LISA} \cite{strubell2018linguistically}. As we can see from the table, {\it DRPC} is significantly better than all the baselines for different amounts of training data, thus demonstrating the better sample complexity of the proposed model for RE.

\begin{table}[t!]
\addtolength{\abovecaptionskip}{-2.0mm}
\addtolength{\belowcaptionskip}{-3mm}
\small
\begin{center}
\begin{tabular}{|l|c|c|c|}
\hline \textbf{System} & \textbf{bc} & \textbf{cts} & \textbf{wl} \\ \hline
CNN & 0.70 & 0.67 & 0.66 \\ \hline
Bi-GRU & 0.69 & 0.66 & 0.64 \\ \hline
C-GCN & 0.73 & 0.76 & 0.72 \\ \hline
LISA & 0.71 & 0.76 & 0.73 \\ \hline
DRPC & {\bf 0.75} & {\bf 0.78} & {\bf 0.74} \\ \hline
\end{tabular}
\end{center}
\caption{\label{similarity} Average cosine similarity between the representations of the relation mentions in the training and test dataset. }
\end{table}

One of the properties we observe in Tables \ref{sota-full}, \ref{sota-noling}, \ref{sota-general} is that the performance of {\it DRPC} and {\it C-GCN} on the {\tt cts} domain is better than those performance on the {\tt bc} and {\tt wl} domains. This is in contrast to the other models where the performance on the {\tt bc} domain is the best, followed by those on {\tt cts} and {\tt wl} \cite{plank2013embedding} (except for {\it LISA} where the performance on {\tt cts} is close to those on {\tt bc}). In order to understand this problem, we run the trained models over the relation mentions in the training and test datasets of ACE 2005 to obtain the final feature representation vectors (e.g., the vectors $o$ in Section \ref{sec:pred}) for the relation mentions. We then compute the average cosine similarity between the pairs of relation mentions where one element comes from the training dataset and the other element belongs to the test dataset. Table \ref{similarity} shows such average cosine similarities for different models over different target domains (i.e., {\tt bc}, {\tt cts} and {\tt wl}). The first observation is that the similarity for {\tt cts} is the largest in {\it DRPC}, {\it C-GCN} and {\it LISA} while this is not the case for the other models. This helps to explain the good performance on the {\tt cts} domain of {\it DRPC} and {\it C-GCN}. Importantly, we also see that the similarities between the target domains and the source domain (i.e., the training data) for {\it DRPC} are better than those for the other methods (esp. {\it CNN} and {\it Bi-GRU}). In other words, {\it DRPC} is able to bridge the gap between domains to achieve better generalization for cross-domain RE, thus also explaining the better operation of {\it DRPC} in this work.

\section{Related Work}

Relation Extraction is one of the main tasks in Information Extraction. Traditional work has mostly used feature engineering with syntactical information for statistical or kernel based classifiers \cite{Zhou:05,bunescu2005shortest,Sun:11,Chan:10}. Recently, deep learning has been introduced to solve this problem with typical architectures such as CNN, LSTM and the attention mechanism \cite{zeng2014relation,Nguyen:15a,zhou2016attention,wang2016relation,nguyen2015combining,zhang2017position,Nguyen:18b}. Using dependency trees in deep learning models has been shown to be effective for RE \cite{xu2015classifying,liu2015dependency,miwa2016end,zhang2018graph}. In this paper, we also use dependency trees to improve RE performance for deep learning; however, we present a novel method to exploit dependency trees where the dependency relations are predicted in a multi-task learning framework for RE. This hasn't been explored in the previous work for RE. 

Cross-domain RE is also a well studied topic. Most of the existing work has investigated genre agnostic features for this setting \cite{plank2013embedding,nguyen2014employing,yu2015combining,gormley2015improved,Nguyen:15b,nguyen2015combining,fu2017domain,Fu:18,shi2018genre}. Our work employs the cross-domain setting as the main evaluation for RE. We demonstrate that decomposing the dependency structures to predict the dependency relations is an effective method to improve the generalization of the models for RE.

Regarding multi-task learning with dependency prediction, the most related work to ours is \cite{strubell2018linguistically} which also predicts the dependency structures in a deep learning model for semantic role labeling. 
Contrary to this work, we predict the existence of a dependency relation between every pair of words in the sentence. The experiments prove that our approach is more effective for RE. 


\section{Conclusion}

In this paper, we introduce a novel model for relation extraction that uses the information in the dependency trees indirectly in a multi-task learning framework. The model jointly predicts dependency relations between words and relations between entity mentions of interest. Moreover, we propose a new control mechanism that regulates the information flow in the model based on the given entity mentions for RE and the gating techniques. The experiments show that the proposed model achieves the state-of-the-art performance for RE on both the general setting and cross-domain setting.

\section*{Acknowledgments} This research is partially supported by the NSF grant CNS-1747798 to the IUCRC Center for Big Learning. 

\bibliographystyle{named}
\bibliography{ijcai19}

\end{document}